\documentclass{INTERSPEECH2023}

\interspeechcameraready

\usepackage[clock]{ifsym}
\usepackage{multirow}
\usepackage{tikz}
\usepackage{pgfplots}
\usepackage{verbatim}
\usepackage{enumitem}
\usepackage{soul}

\title{Investigating Pre-trained Audio Encoders in the Low-Resource Condition}

\name{Hao Yang, Jinming Zhao, Gholamreza Haffari, Ehsan Shareghi}
\address{
  Department of Data Science \& AI, Monash University}
\email{first.last@monash.edu}

\begin{document}

\maketitle
 
\begin{abstract}
Pre-trained speech encoders have been central to pushing state-of-the-art results across various speech understanding and generation tasks. 
Nonetheless, the capabilities of these encoders in low-resource settings are yet to be thoroughly explored. To address this, we conduct a comprehensive set of experiments using a representative set of 3 state-of-the-art encoders (Wav2vec2, WavLM, Whisper) in the low-resource setting across 7 speech understanding and generation tasks. 
We provide various quantitative and qualitative analyses on task performance, convergence speed, and representational properties of the encoders. 
We observe a connection between the pre-training protocols of these encoders and the way in which they capture information in their internal layers. In particular, we observe the Whisper encoder exhibits the greatest low-resource capabilities on content-driven tasks in terms of performance and convergence speed.\footnote{https://github.com/YangHao97/investigateAudioEncoders} 


\end{abstract}
\noindent\textbf{Index Terms}: speech encoders, low-resource setting, speech understanding

\section{Introduction}

In recent years, the advancement in various speech tasks has largely been driven by encoder models that are typically pre-trained on large-scale datasets via self-supervised learning~\cite{mohamed2022self,DBLP:journals/patterns/LiuMPQJKHS22}. As the prominent examples, the Wav2vec2~\cite{baevski2020wav2vec} model leverages a speech quantiser module to simulate token prediction of BERT~\cite{kenton2019bert}, while  HuBERT~\cite{hsu2021hubert} adopts a  clustering method to produce discrete labels for each 
feature vector to imitate masked language model loss. Similar to HuBERT, WavLM~\cite{chen2022wavlm} proposes a denoising masked speech modelling, which masks segments of speech signals to predict the pseudo-label at the output. 

%
While it is expected that these pre-trained encoders produce universal speech features effective for a broad range of downstream tasks, in practice pre-trained models still require  large amounts of fine-tuning labelled data to produce state-of-the-art performance, or to converge. This could be attributed to their inefficiency in utilising the representation space~\cite{DBLP:conf/emnlp/YangZHS22}, as well as the difference between the objectives for pre-training and fine-tuning steps~\cite{radford2022robust}. For instance, the pre-training objective is typically designed in the absence of any textual or content cue (i.e., to predict masked speech segments), while the downstream tasks (i.e., automatic speech recognition and speech translation) often require a mapping between speech and text. 
%
%
%
%
An exception in this space is the Whisper encoder-decoder model~\cite{radford2022robust}, which leverages weak supervision through large scale crawled data of~(audio, transcript) pairs from the internet, and is pre-trained by learning the mapping between speech and decoder outputs (i.e., in transcription or and translation).
%
%
%

To better understand the interplay between pre-training protocols of speech encoders, the amount of fine-tuning data, and speech task types, we conduct a comprehensive study in this work. We evaluate a set of three very recent speech models (Wav2vec2, WavLM, and Whisper) and assess their performance on 7 downstream tasks (covering content, speaker and semantic types) in the low-resource setting. 
Through extensive experiments in the low-resource setting, we found that  Whisper significantly outperforms Wav2vec2 and WavLM by a large margin on content-related (content, semantics) tasks, and shows performance degradation when speaker information is required for a downstream task. To investigate how this behaviour is connected with Whisper's pre-training and representational properties, 
we examine layer-wise information of Whisper and the other baselines. Additionally, through qualitative and quantitative analyses, we highlight how Whisper's superior performance could be attributed to the properties of its representational space. 
%
{We hope our study to provide insights for a more effective use of pre-trained speech encoders in the resource-constrained setting.}



\begin{table*}[t]
\caption{Main results and the number of updates~\VarClock~required for fine-tuning in low-resource scenarios. The encoders' details are as follows: W2V2: 317M/24 layers, WavLM: 317M/24 layers, Whisper \textsc{Base}: 21M/6 layers, Whisper \textsc{Small}: 88M/12 layers, Whisper \textsc{Medium}: 307M/24 layers. The \textbf{bold} font and \underline{underlined} numbers denote the fastest convergence speed and performance, respectively.}
\label{table:main}
\setlength{\tabcolsep}{6pt} 
\centering
\scalebox{0.85}{ 
\begin{tabular}{llcccccccccccccc}
\toprule
 & & \multicolumn{2}{c}{SD}  & \multicolumn{2}{c}{SF}  &  \multicolumn{2}{c}{IC} & \multicolumn{2}{c}{KS} & \multicolumn{2}{c}{ASR} & \multicolumn{2}{c}{SID} & \multicolumn{2}{c}{ST} \\
 \cmidrule(lr){3-4}\cmidrule(lr){5-6}\cmidrule(lr){7-8}\cmidrule(lr){9-10}\cmidrule(lr){11-12}\cmidrule(lr){13-14}\cmidrule(lr){15-16}
 Tr.&Model& DER $\downarrow$  &\VarClock& F1 $\uparrow$ &\VarClock  & Acc $\uparrow$ &\VarClock & Acc $\uparrow$ &\VarClock &  WER $\downarrow$&\VarClock & Acc $\uparrow$&\VarClock & BLEU $\uparrow$&\VarClock\\
 \midrule 
 \bf \parbox[t]{0mm}{\multirow{5}{*}{\rotatebox[origin=c]{90}{$1\%$}}}
 &\text{W2V2}&10.23 & 6.7k &57.88 & 42k& 12.54 & 6.8k & 85.17&9.2k &  99.99 & 50k & 9.74 & 4.5k & 0.17 & \textbf{3k}\\
 &\text{WavLM}&6.38 &0.4k& 75.56 &98k& 26.02 &3.25k& 93.57&10k& 17.84&8.4k& \underline{12.69} & 11k & 0.69 & 14k\\
 &Whisper-\text{\textsc{Base}}&7.24 & \textbf{0.2k} &70.47 & 84k& 67.04 & \textbf{2.75k} & \underline{96.79}& 1k &  26.43 & 16k & 2.66 & \textbf{2k} & 0.87 & 30k\\
 &Whisper-\textsc{Small}&  5.37 &1.2k&74.45 &\textbf{38k} & 57.63 & 4.25k &96.62& \textbf{0.5k}& 20.27& 10k & 3.35 & 3.5k & 0.94 & 28k\\
 &Whisper-\textsc{Medium}&  \underline{5.23} & 0.4k &\underline{77.76} &48k & \underline{73.74} &3k&96.72& 0.75k & \underline{17.56} & \textbf{5k} & 3.97 & 4.5k & \underline{0.98} & 24k\\
 
\midrule
\bf \parbox[t]{0mm}{\multirow{5}{*}{\rotatebox[origin=c]{90}{$5\%$}}} 
&\text{W2V2}&9.20 &4k  &78.29 &\textbf{58k}& 53.07 &16k&94.25&20k & 14.70 & 100k & 41.90 & 47k & 0.20 & \textbf{4k}\\
 &\text{WavLM}&5.16 &1.8k &86.50 &92k& 91.30 &5k&95.91&\textbf{1k} & \underline{7.90} & \textbf{50k} & \underline{55.52} & 27k & 4.19 & 12k\\
 &Whisper-\textsc{Base}&6.84 &1.6k &82.80 &94k& 95.39 &3k&97.44&\textbf{1k} & 16.18 & 100k & 11.63 & 15k & 3.41 & 20k\\
 &Whisper-\textsc{Small}&4.89 &\textbf{1k} &85.83 &70k& 95.78 &\textbf{2.5k}&97.73&\textbf{1k} & 11.76 & 90k & 13.47 & 16k & 3.84 & 26k\\
 &Whisper-\textsc{Medium}&\underline{4.59} &2.4k &\underline{87.60} &62k& \underline{98.23} &\textbf{2.5k}&\underline{97.95}&\textbf{1k} & 9.75 & 84k & 17.94 & \textbf{13k} & \underline{4.22} & 30k\\
 
  \midrule
   \bf \parbox[t]{0mm}{\multirow{5}{*}{\rotatebox[origin=c]{90}{$10\%$}}} 
&\text{W2V2}&8.21 &6k &80.74 & 90k& 77.91 &45k & 95.85&15.5k& \underline{5.96} &90k & 56.09 & 78k & \underline{7.21} & 25k\\
 &\text{WavLM}&4.76 &1k &88.84 &\textbf{80k}& 94.38 &\textbf{2.5k}&96.82&\textbf{0.5k} & 5.99 & 98k & \underline{79.51} & 61k & 6.99 & \textbf{22k}\\
 &Whisper-\textsc{Base}&5.89 &\textbf{0.2k} &85.15 &86k& 96.92 &3k&97.24&3k & 13.41 & 100k & 19.48 & \textbf{13k} & 5.19 & 28k\\
 &Whisper-textsc{Small}&4.69 &0.6k &87.90 &98k& 96.44 &\textbf{2.5k}&97.63&2k & 9.47 & 86k & 23.04 & \textbf{13k} & 6.09 & 23k\\
 &Whisper-\textsc{Medium}&\underline{4.38} &0.4k &\underline{89.80} &96k& \underline{98.78} &7.5k&\underline{97.96}&1k & 7.74 & \textbf{74k} & 30.05 & \textbf{13k} & 6.48 & 29k\\
 
\bottomrule
\end{tabular}
}
\end{table*}
\section{Related work}\label{sec:relwork}
We provide a brief overview of some of the well-known pre-trained speech models. The Wav2vec~\cite{schneider19_interspeech} model proposed two multi-layer convolutional neural networks stacked on top of each other to map raw audio to a representation instead of traditional acoustic feature extraction. Subsequently, Wav2vec 2.0~\cite{baevski2020wav2vec} attached Transformer layer~\cite{vaswani2017attention} to the feature extractor layer and utilised InfoNCE loss~\cite{oord2018representation} and quantiser modules to predict masked spans of the representation at the output. Similar to BERT~\cite{kenton2019bert} in the text domain, HuBERT~\cite{hsu2021hubert} adopts the clustering method to produce discrete labels for each input feature, with the purpose of imitating masked language model loss. WavLM~\cite{chen2022wavlm} is proposed with denoising masked speech modeling, which randomly transforms the input audio and masks 50\% of speech signals to predict the labels corresponding to the masked positions. Additionally, it follows the idea proposed by HuBERT~\cite{hsu2021hubert}, converting continuous signals into discrete labels through a clustering method, and models the discrete labels as targets. WavLM achieves state-of-art results on several downstream tasks from the SUPERB benchmark~\cite{DBLP:conf/interspeech/YangCCLLLLSCLHT21}. As an exception to the above models, Whisper~\cite{radford2022robust} is pre-trained under weak supervision through crawled audio-transcript pairs from the internet. Transcription and translation are set as pre-training targets, pre-training the model via learning to map the input audio to its transcript as the output.

Hsu et al.~\cite{hsu21_interspeech} highlighted the benefits of pre-training on several domains. 
The pre-trained audio representations have been investigated from different aspects~\cite{9688137, shah2021all,pasad2021layer,chung2021similarity,pasad2022comparative,DBLP:conf/interspeech/ZhaoYHS22}, indicating they can generalise to wide range of corpora.
However, Yang et al.~\cite{DBLP:conf/emnlp/YangZHS22} demonstrated that the Wav2vec2 speech encoder under-utilises the representation space, and proposed a self-supervision approach to improve the representation isotropy, leading to faster convergence during downstream task training. 
%
Yi et al.~\cite{9533587} applied Wav2vec2 in low-resource conditions for multilingual speech recognition and verified the potential for transfer-ability of monolingual Wav2vec2 to other languages. The Whisper encoder-decoder model has exhibited its capabilities in zero-shot settings~\cite{radford2022robust} by achieving state-of-art performance on various tasks, from multilingual ASR, and translation, to Language Identification, and Long-form Transcription.

\begin{figure*}[t]
    \centering
    \parbox[c]{5mm}{KS} 
    \parbox[c]{31mm}{\includegraphics[trim={6cm 3.3cm 4.5cm 3.5cm},clip, scale=0.105]{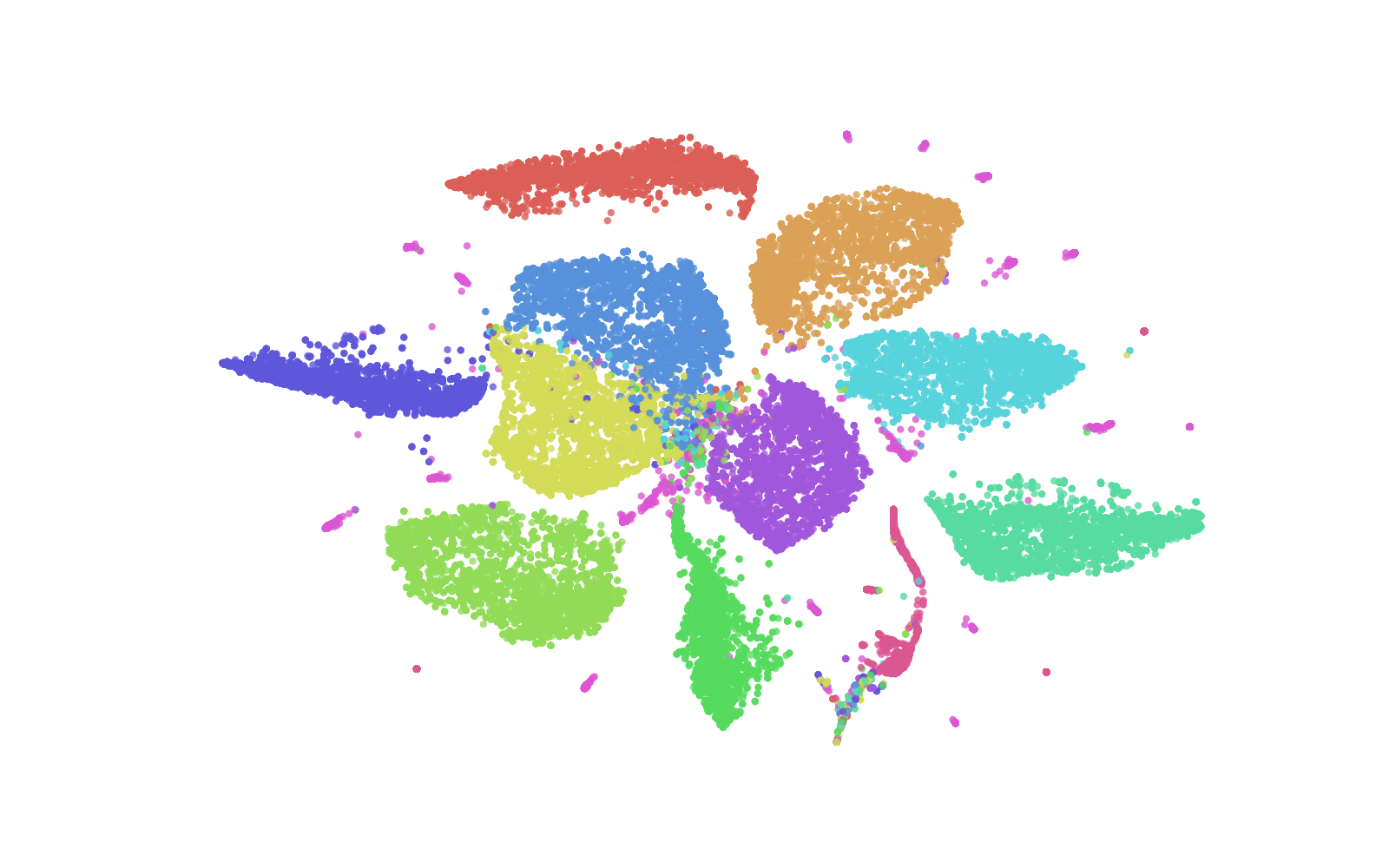}} 
    \parbox[c]{31mm}{\includegraphics[trim={6cm 3.3cm 4.5cm 3.5cm},clip, scale=0.105]{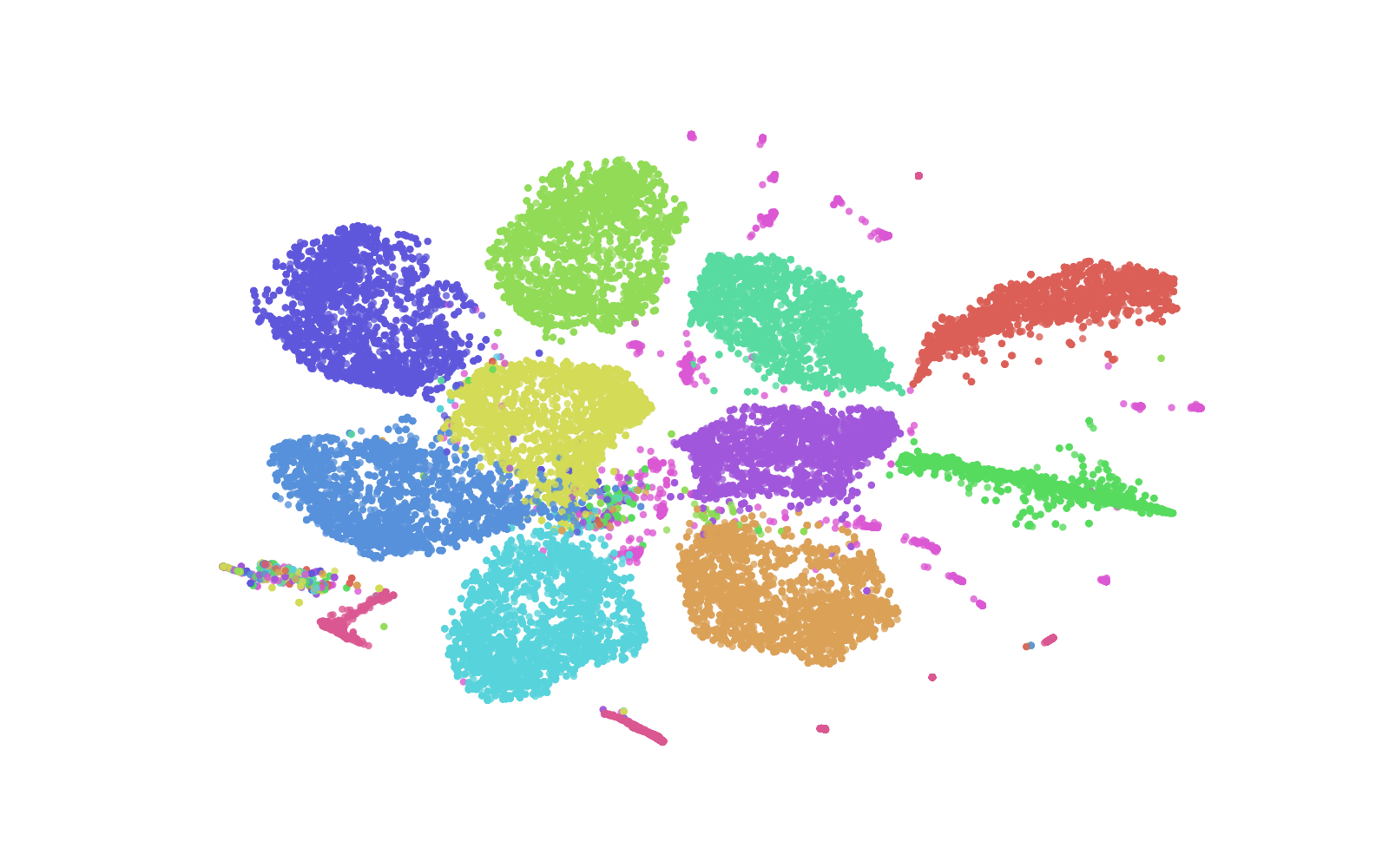}} 
    \parbox[c]{31mm}{\includegraphics[trim={6cm 3.3cm 4.5cm 3.5cm},clip, scale=0.105]{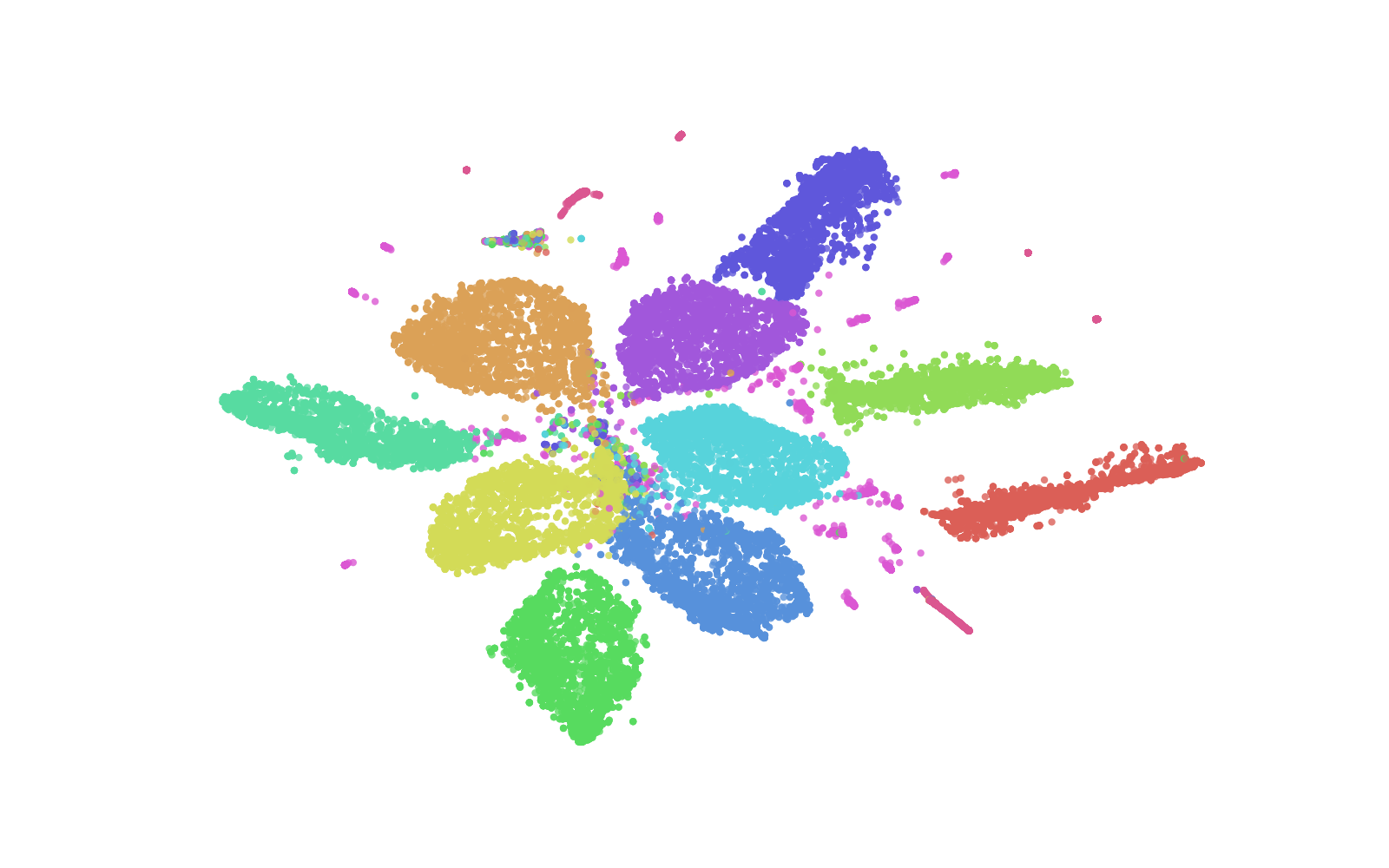}}
    \parbox[c]{31mm}{\includegraphics[trim={6cm 3.3cm 4.5cm 3.5cm},clip, scale=0.105]{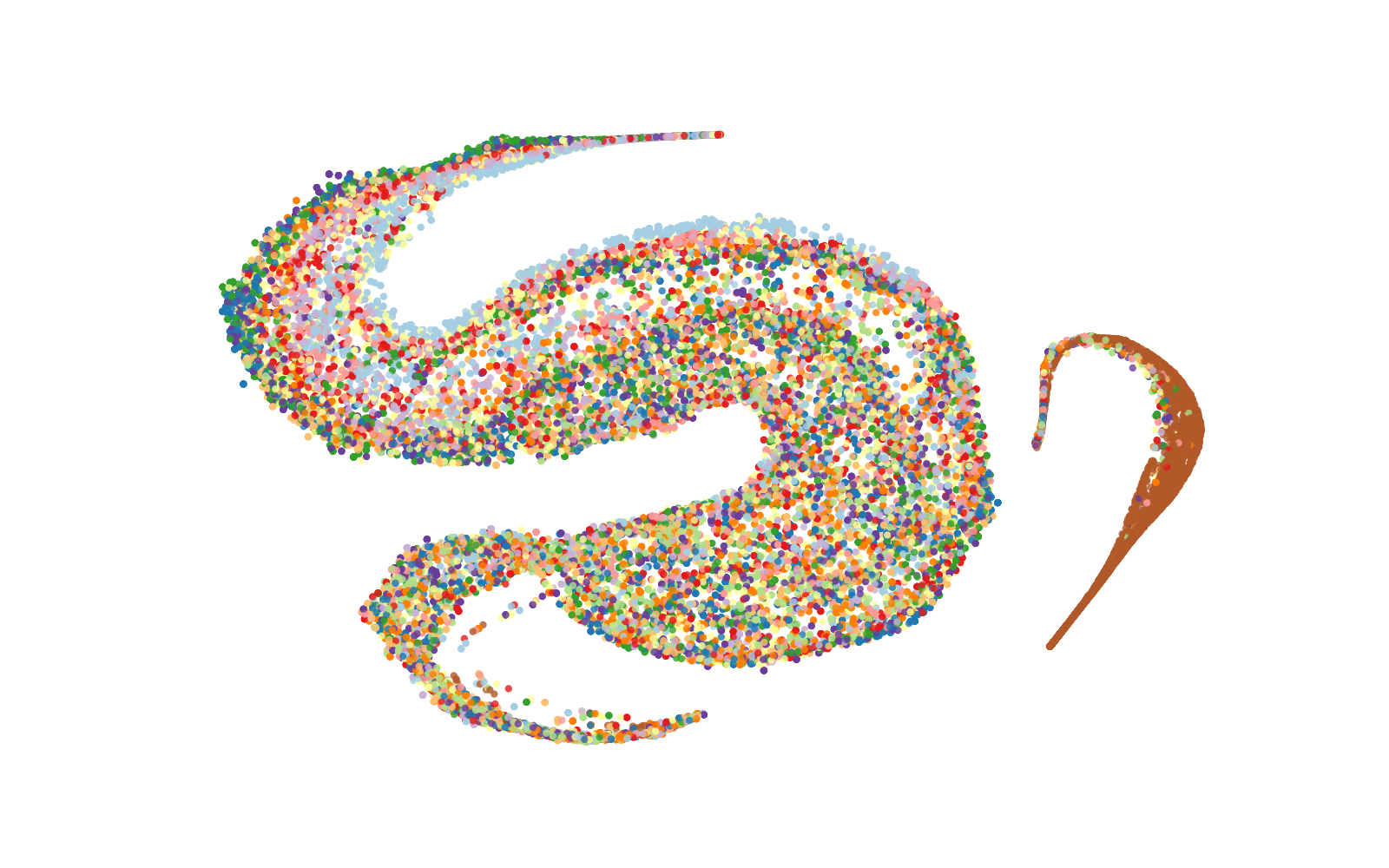}}
    \parbox[c]{31mm}{\includegraphics[trim={6cm 3.3cm 4.5cm 3.5cm},clip, scale=0.105]{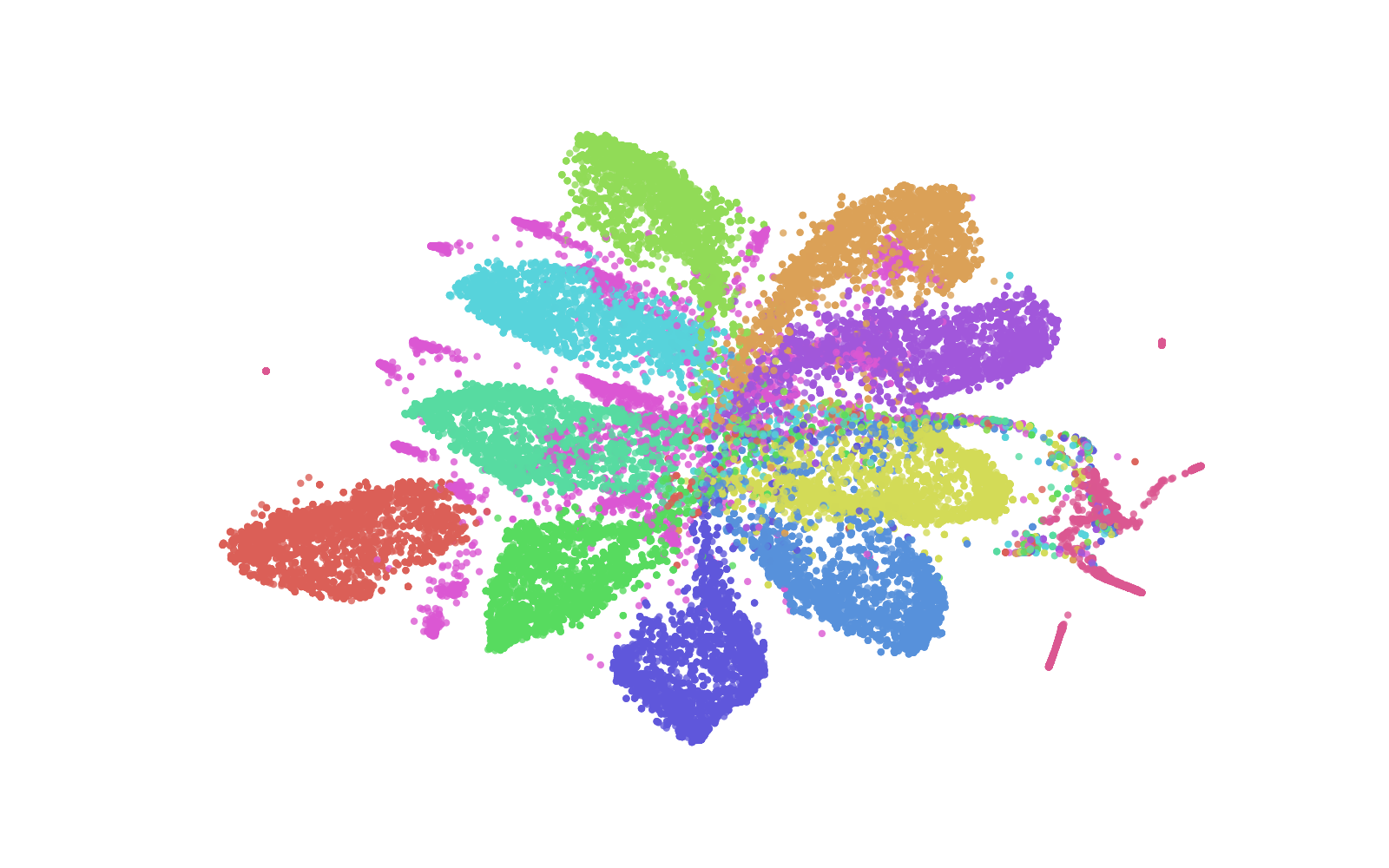}}\\
    \parbox[c]{5mm}{IC} 
    \parbox[c]{31mm}{\includegraphics[trim={6cm 3.3cm 4.5cm 3.5cm},clip, scale=0.105]{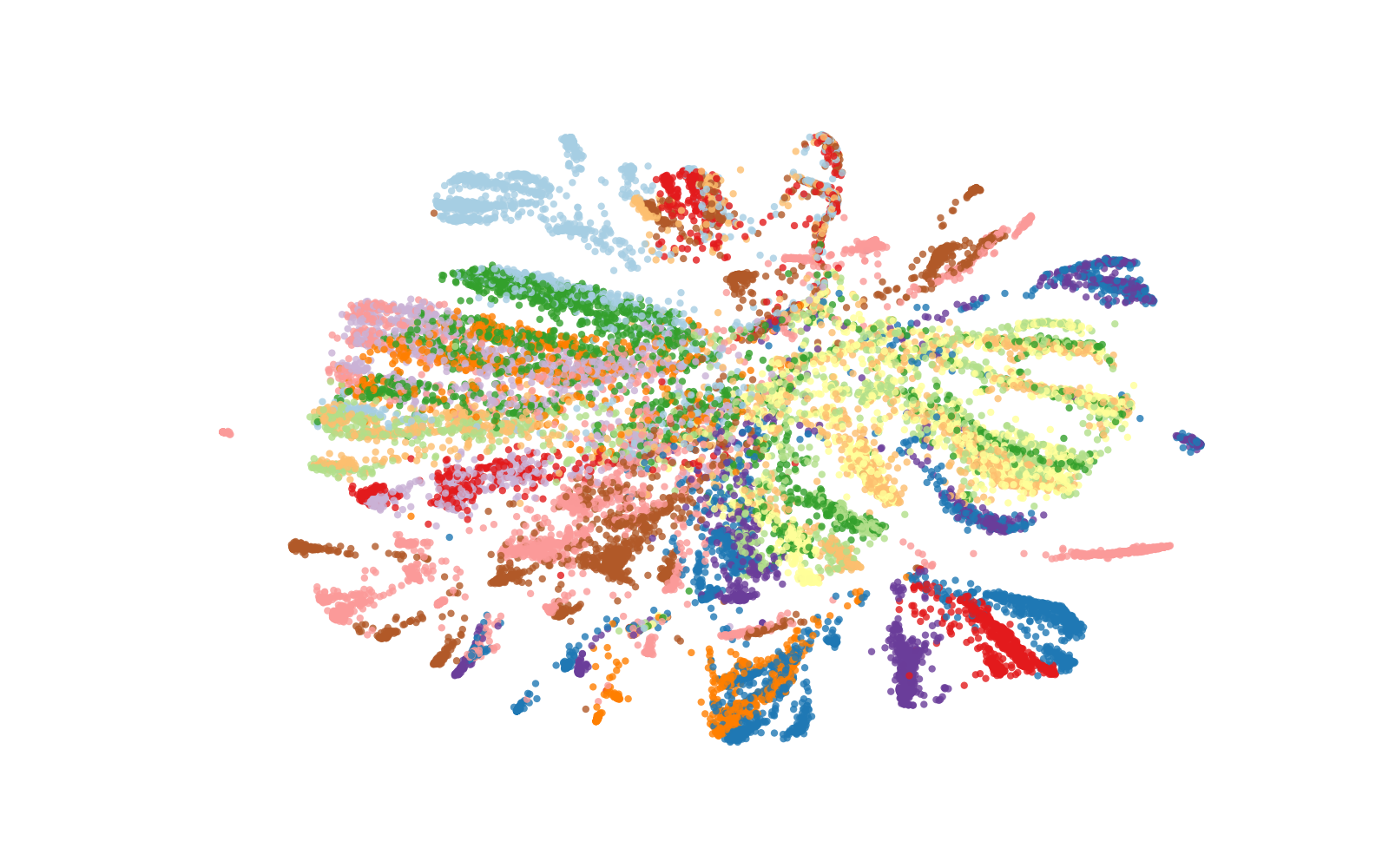}} 
    \parbox[c]{31mm}{\includegraphics[trim={6cm 3.3cm 4.5cm 3.5cm},clip, scale=0.105]{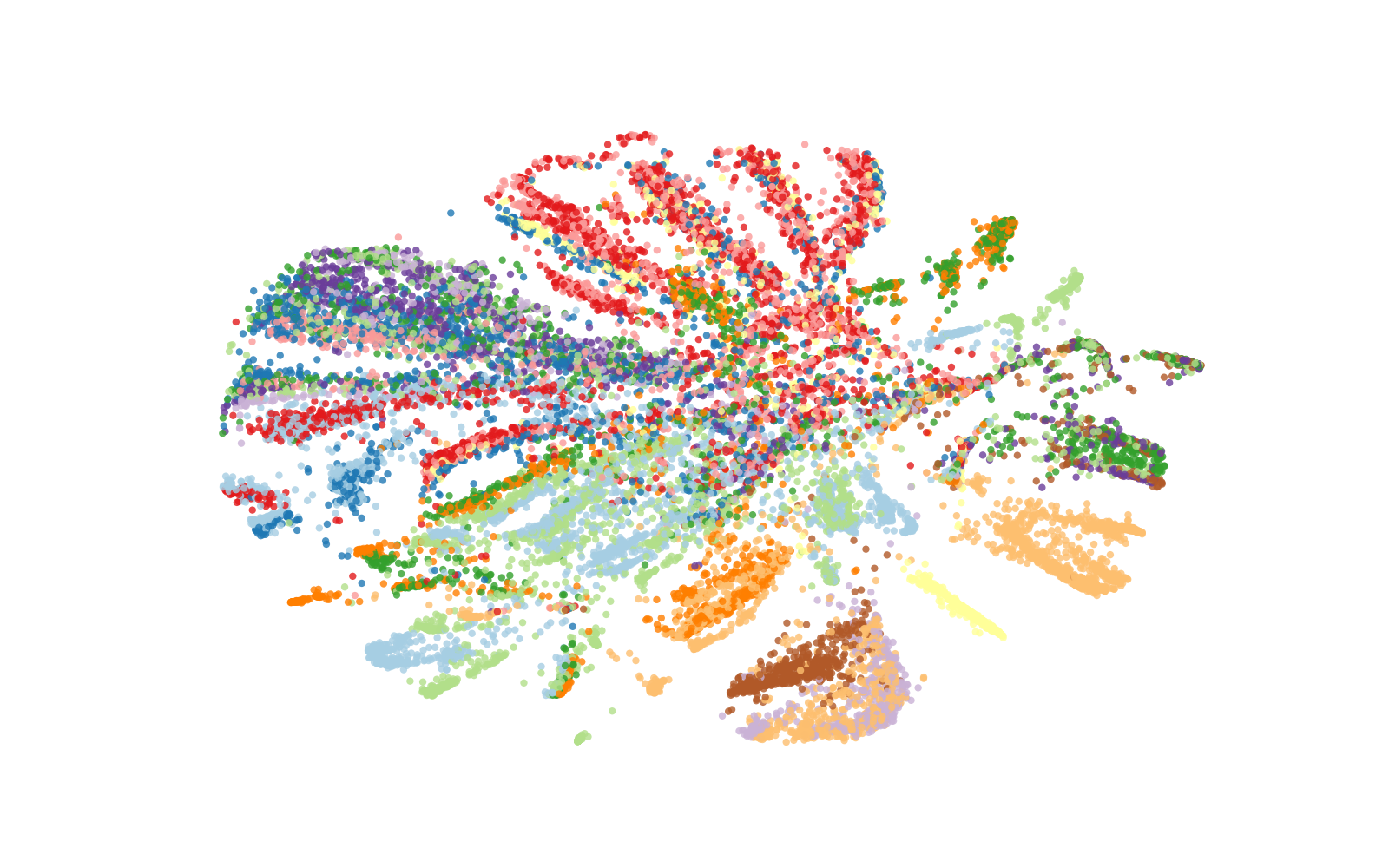}} 
    \parbox[c]{31mm}{\includegraphics[trim={6cm 3.3cm 4.5cm 3.5cm},clip, scale=0.105]{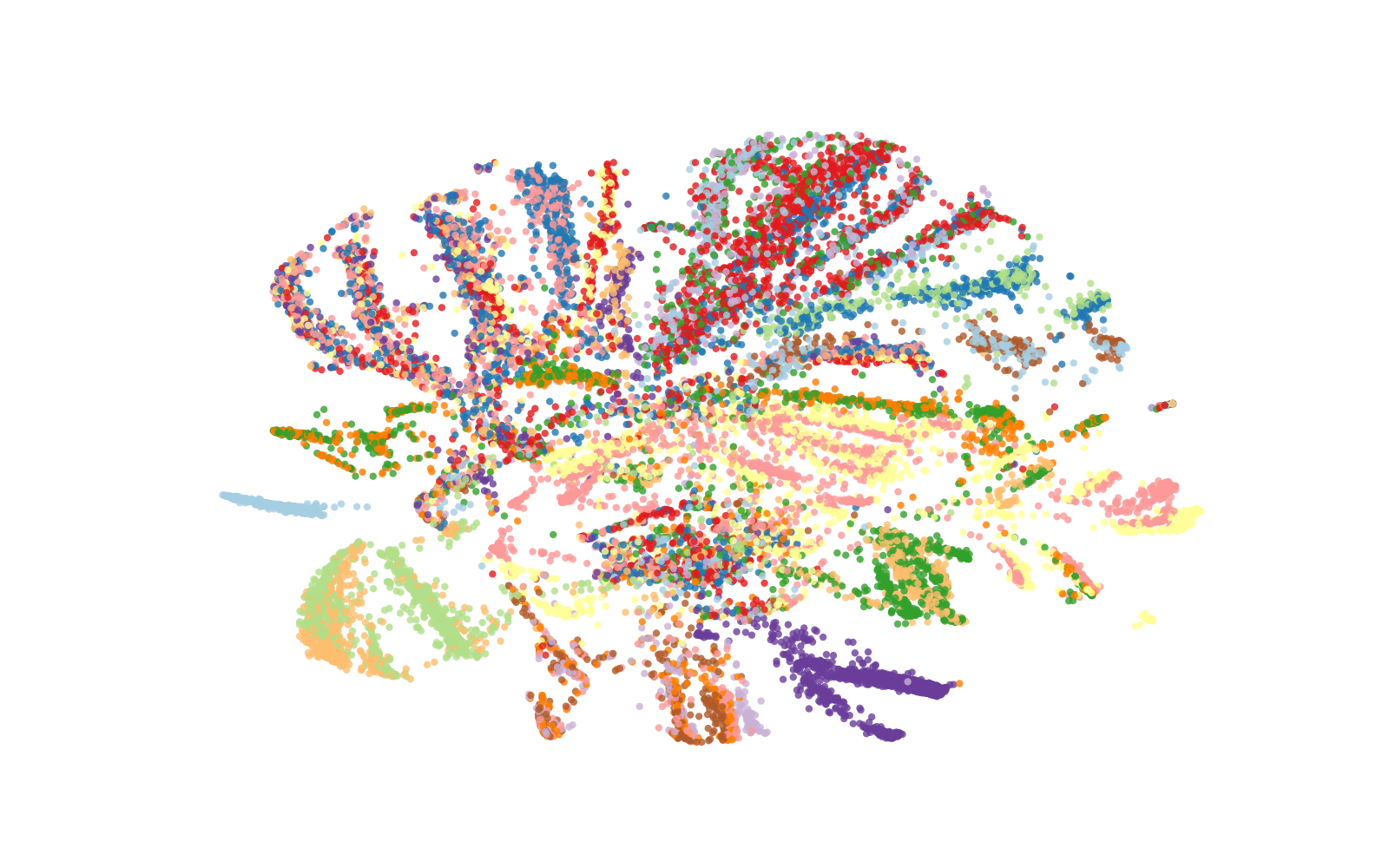}}
    \parbox[c]{31mm}{\includegraphics[trim={6cm 3.3cm 4.5cm 3.5cm},clip, scale=0.105]{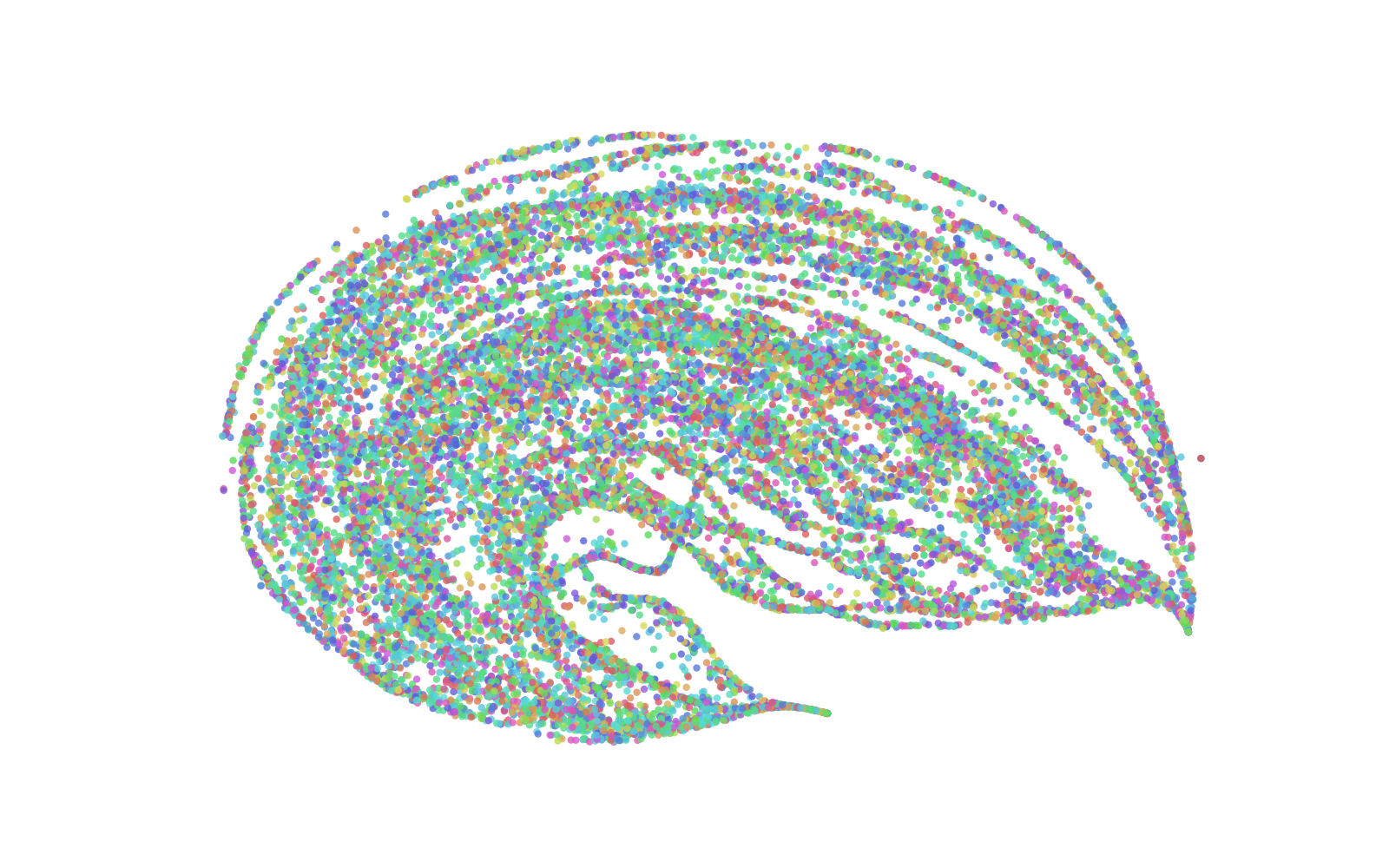}}
    \parbox[c]{31mm}{\includegraphics[trim={6cm 3.3cm 4.5cm 3.5cm},clip, scale=0.105]{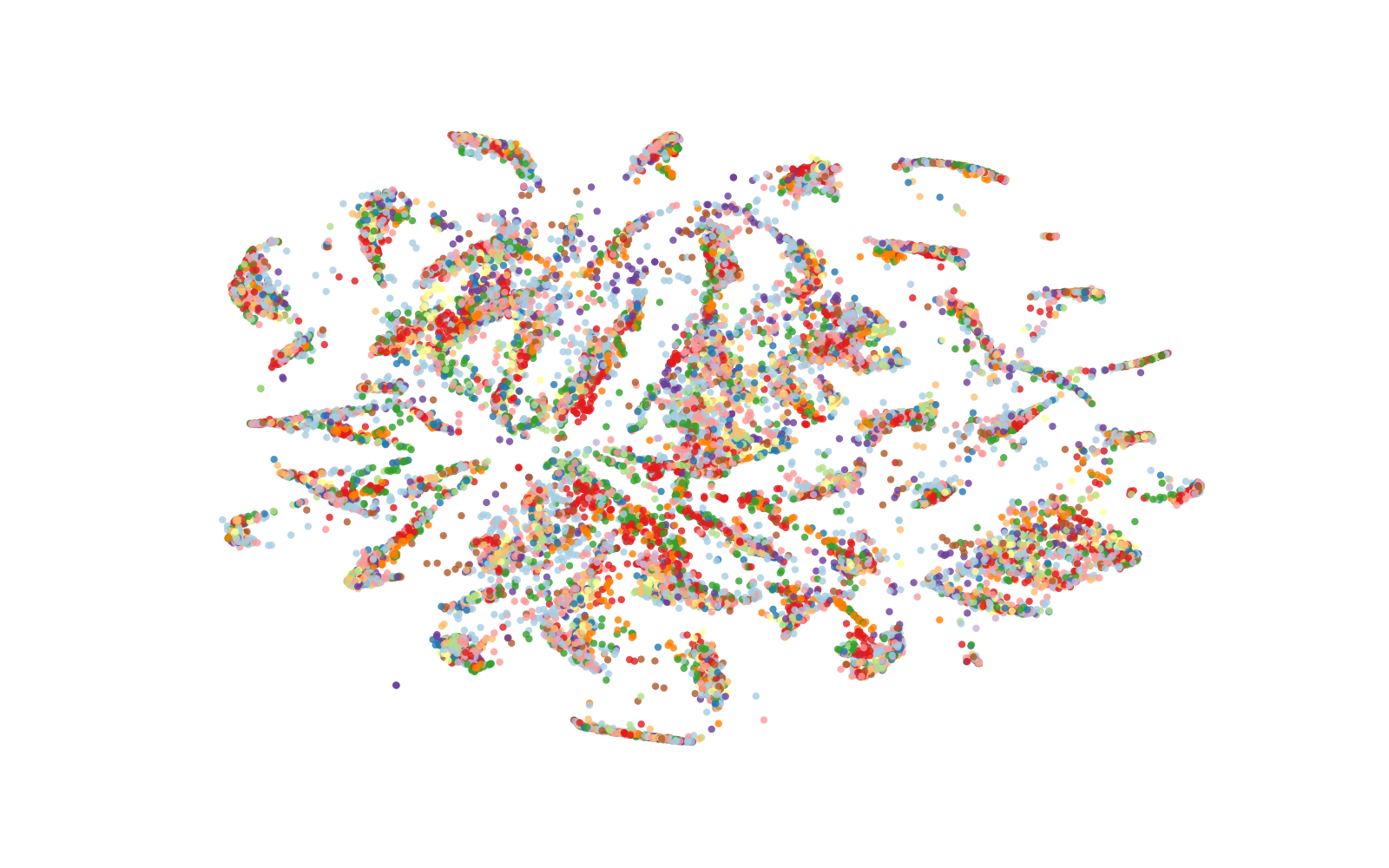}}\\
    \begin{tabular}{l c c c c c c c c }
    &&&&\parbox[c][6mm]{27.4mm}{\textsc{Base}} &
    \parbox[c][6mm]{27.4mm}{\textsc{Small}} &
    \parbox[c][6mm]{27.4mm}{\textsc{Medium}} &
    \parbox[c][6mm]{27.4mm}{W2V2} &
    \parbox[c][6mm]{27.4mm}{WavLM} 
    \end{tabular}
    \caption{t-SNE visualisation of the representation spaces produced by the encoders on KS (top) and IC (bottom) tasks training set (prior to fine-tuning) with colours indicating class labels. 
    }
    \label{fig:tsne}
\end{figure*}
\section{Experiments}
In this section, we first describe our experimental settings (\S\ref{sec:expset}). Next, we report the results on 7 downstream tasks in low-resource scenarios (\S\ref{sec:main_results}). Lastly, we provide an analysis of {Whisper} encoders on the quantitative and qualitative properties compared to two other widely used encoders, Wav2vec2 and WavLM (\S\ref{sec:analysis_discuss}).

\subsection{Experimental Settings}\label{sec:expset}
\noindent\textbf{Tasks and Dataset.} We conducted experiments on various tasks from SUPERB and SUPERB-SG benchmarks:\footnote{https://superbbenchmark.org} Automatic Speech Recognition (ASR), Speaker Diarisation (SD), Intent Classification (IC), Slot Filling (SF), Keyword Spotting (KS), Speaker Identification (SID), and Speech Translation (ST). For evaluation, we use word error rate (WER), diarisation error rate (DER),  accuracy (ACC), slot-type F1 score, accuracy (ACC), accuracy (ACC), and BLEU score, respectively. 
To simulate training in the low-resource setting, for a given task we randomly sample 1\%, 5\% and 10\% from the corresponding training set. The statistics of these data splits are reported in Table~\ref{tab:data}. 

\noindent\textbf{Models.} We use three versions of {Whisper} encoders\footnote{{For brevity, we drop encoder when Whisper is mentioned.}}, including base.en, small.en and medium.en\footnote{{https://github.com/openai/whisper}}, denoted as  \textsc{Base}, \textsc{Small} and \textsc{Medium}. Our baseline models are \textsc{Wav2vec 2.0 Large\footnote{{https://huggingface.co/facebook/wav2vec2-large-lv60}}} (W2V2)~\cite{baevski2020wav2vec} and \textsc{WavLM Large\footnote{{https://huggingface.co/microsoft/wavlm-large}}} (WavLM)~\cite{chen2022wavlm}.
We report {the maximum} number for training updates  in Table \ref{tab:data}. We use the SUPERB evaluation pipeline by \emph{freezing} the encoders for downstream tasks while attaching a benchmark-specified lightweight prediction head for each task, unless mentioned otherwise. {We adopt the identical training configuration (e.g., batch size, optimizer) for all models based on SUPERB hyperparameter settings.} {Experiments were done on 1xRTX 6000 GPU with 48GB Memory.}
\begin{table}[t]
\caption{Training tasks' types and splits, and the corresponding training data sizes / {cap on} training updates.}
\label{tab:data}
\small
    \centering
    \scalebox{0.9}{
    \begin{tabular}{l c c c c c c c}
        \toprule
         \cmidrule(lr){2-7}
         Task& Type    &\multicolumn{2}{c}{1\%}&\multicolumn{2}{c}{5\%}&\multicolumn{2}{c}{10\%}\\
         \cmidrule(lr){1-1}\cmidrule(lr){2-2}\cmidrule(lr){3-4}\cmidrule(lr){5-6}\cmidrule(lr){7-8}
    SD  & speaker &   \multicolumn{2}{c}{0.14k / 20k} & \multicolumn{2}{c}{0.70k / 20k} & \multicolumn{2}{c}{1.39k / 50k}      \\
    SID  &speaker&   \multicolumn{2}{c}{1.38k / 20k} & \multicolumn{2}{c}{6.92k / 50k} & \multicolumn{2}{c}{13.8k / 100k}      \\
    SF  & semantics&  \multicolumn{2}{c}{1.05k / 100k} & \multicolumn{2}{c}{5.24k / 100k} & \multicolumn{2}{c}{10.5k / 100k} \\
    IC &semantics& \multicolumn{2}{c}{0.23k / 20k} & \multicolumn{2}{c}{1.16k / 20k} & \multicolumn{2}{c}{2.32k / 50k} \\
    KS &content& \multicolumn{2}{c}{0.51k / 20k} & \multicolumn{2}{c}{2.56k / 50k} & \multicolumn{2}{c}{5.11k / 50k}\\
    ASR  &content& \multicolumn{2}{c}{0.28k / 50k} & \multicolumn{2}{c}{1.43k / 100k} & \multicolumn{2}{c}{2.86k / 200k}\\
    ST   &semantics& \multicolumn{2}{c}{2.88k / 32k} & \multicolumn{2}{c}{14.4k / 32k} & \multicolumn{2}{c}{28.8k / 32k}\\
    \bottomrule
    \end{tabular}
}

\end{table}
\subsection{Main Results}\label{sec:main_results}
We report results in Table~\ref{table:main}.
Overall, {Whisper} variants outperform W2V2 and WavLM on the majority of tasks in various data conditions with fewer updates except for SID. We summarise the findings for each task as follows:
\setlist[itemize]{align=parleft,left=0pt..1em}
\begin{itemize}
\item \textbf{IC} 
Various {Whisper} models exhibit a significantly better performance in all settings. Even with 1\% of fine-tuning data, \textsc{Base} surpasses WavLM by 150\% with a faster convergence rate. As the size of training data increases, {Whisper} on average converges 10$\times$ {faster} than W2V2.
\item \textbf{SF} As data resources become more scarce, the benefits of using {Whisper} become more eminent in training speed and performance. Notably, \textsc{Medium} outperforms the baselines by a large margin with 50\% less number of updates. 
\item \textbf{SD} 
\textsc{Medium} significantly outperforms the baselines in all settings. \textsc{Base} and \textsc{Small} (despite being 75\% and 90\% smaller) converge much faster with better or comparable performance compared to W2V2 and WavLM. Note that although SD is regarded as a speaker task, content information is still required as models tend to distinguish speaker timestamps by content, not just speaker features.

\item \textbf{KS} {Whisper} dramatically boosts the performance with higher convergence speed. \textsc{Base} and \textsc{Medium}, fine-tuned on 1\% and 5\% of task data, surpass W2V2 (96.66) and WavLM (97.86) models that are fine-tuned on 100\% of task data.\footnote{The numbers are obtained from SUPERB Leaderboard at the time of writing this paper.} 

\item \textbf{ASR} {Whisper} models achieve robust performance at 1\%, i.e., the extremely low data condition, with a few thousand updates. \textsc{Medium}  outperforms WavLM even though the latter was pre-trained on Librispeech and the former was not, whereas W2V2 has difficulties to converge. W2V2 and WavLM gradually pick up as the amount of training instances increases.

\item \textbf{SID} {Whisper} models perform poorly on this task. Our hypothesis is that Whisper pre-training places emphasis on capturing content (via mapping audios to text) rather than speaker information and speech features that are important in SID (a speaker task). In contrast, W2V2 and WavLM, which are pre-trained only on speeches, are better positioned to tackle this task. We will unpack this hypothesis later.
\item \textbf{ST} {Whisper} models do not perform well on translation, even with the increase in training corpus size. This could be due to replacing the internal Whisper decoder (i.e., used during the pre-training phase) with SUPERB's decoder. Nonetheless, {Whisper} still achieves the best performance at 1\% and 5\% compared to the baselines. 
\end{itemize}

\subsection{Analysis and Discussion}\label{sec:analysis_discuss}
In this section, we start with a qualitative comparison of the pre-trained representations produced by {Whisper} variants, W2V2 and WavLM. We then measure the utilisation of the representation space through isotropy, and finish by investigating the information captured at different layers of these encoders.

\noindent\textbf{t-SNE.} 
We create t-SNE visualisations of the training data of KS and IC with the vanilla encoders, as shown in Figure \ref{fig:tsne}. On KS, the embedding space of {Whisper} 
exhibits a better clustering of speech representations compared to W2V2 and WavLM, facilitating a much faster 
fine-tuning convergence and better task performance. This  is less eminent on IC, although representations of {Whisper} are still better clustered than the baselines. This also explains why the performance of {Whisper} at 1\% on IC is less remarkable compared with the KS task (but still far exceeds the baselines). Furthermore, the relatively tangled embedding spaces are partially due to the IC task having more classes than KS (31 vs. 12), which increases the overall task difficulty. We also produced the visualisation on SID, and observed a much worse clustering pattern compared with IC, explaining the weaker performance of {Whisper} on this task.
\newline
\noindent\textbf{Isotropy.} 
The geometry of representations generated by pre-trained Transformer have been shown to suffer from the anisotropy problem~\cite{arora2016latent}. Ideally the representations should be uniformly distributed in a spherical space~(isotropic), but in practice they only occupy narrow regions of the embedding space (anisotropic)~\cite{li2020sentence}. {The average isotropy scores of W2V2, WavLM, and {Whisper} (average of the three versions) on 7 downstream task datasets} are 1e-300, 1e-14, and 1e-2, respectively.
While being several orders of magnitude better than the other baselines, the Whisper's isotropy even approaches that of a text embedding space (i.e., compared with 1e-1 of the MirrorBert text encoder~\cite{liu2021fast}). {This indicates that Whisper largely mitigates the 
anisotropic problem that the other baselines face.}
\begin{figure}[!h]
\centering

\begin{tikzpicture}[thick,scale=0.78, every node/.style={scale=1}]
    \begin{axis}[
        legend style={at={(0.12,0.71)},anchor=center},
        title style={at={(0.5,0.97)},anchor=north, yshift=-0.1, draw=gray},
        xlabel=Layers (W2V2),
        ylabel=Weights,
        ymajorgrids, tick align=inside,
        major grid style=dashed,
        height=5cm, 
        width=10cm,
        xmin=0, xmax=25,
        ymin=0, ymax=1.0,
        ylabel near ticks,
        ytick={      0.1,  0.2, 0.3,  0.4,  0.5, 0.6, 0.7, 0.8, 0.9, 1.0, 1.1},
        yticklabels={0.05, 0.1, 0.15, 0.2, $//$, 0.7, 0.8, 0.9, 1.0}
        ]
    \addplot[smooth,color=blue] plot coordinates {
(0, 0.014477184042334557)
(1, 0.020635377615690224)
(2, 0.021352179348688775)
(3, 0.021224291997516731)
(4, 0.02159113623201847)
(5, 0.026323683559895616)
(6, 0.029665244743227957)
(7, 0.0361102856690552)
(8, 0.04547366127317189)
(9, 0.050607040524482724)
(10, 0.056328803300857505)
(11, 0.07438330323607518)
(12, 0.11065570265054704)
(13, 0.1013345792889595)
(14, 0.15575236082207026)
(15, 0.16521514793882097)
(16, 0.20166213381043878)
(17, 0.23989080872867584)
(18, 0.1670480089426104)
(19, 0.1683707112083435)
(20, 0.12884413209275034)
(21, 0.12669926880733217)
(22, 0.006440576327115883)
(23, 0.003722786883658114)
(24, 0.005662190792885935)
    };
    \addlegendentry{IC}

    \addplot[smooth,color=red]
        plot coordinates {
(0, 0.034296661615371704)
(1, 0.04860270395874977)
(2, 0.05047047883272171)
(3, 0.05507121235132217)
(4, 0.0546843484044075)
(5, 0.05948226153850555)
(6, 0.06941913068294525)
(7, 0.07102202624082566)
(8, 0.07584554702043534)
(9, 0.08556811511516571)
(10, 0.08400307595729828)
(11, 0.08798979967832565)
(12, 0.09412646293640137)
(13, 0.08524616807699204)
(14, 0.10275176167488097)
(15, 0.12203816324472428)
(16, 0.13108700513839722)
(17, 0.13341882824897767)
(18, 0.11274378001689911)
(19, 0.11868075281381607)
(20, 0.13523893058300017)
(21, 0.12694406509399414)
(22, 0.019013594835996627)
(23, 0.015816457569599152)
(24, 0.026438694447278976)
        };
    \addlegendentry{KS}

    \addplot[smooth,color=cyan]
        plot coordinates {
(0, 0.028069650754332542)
(1, 0.03675244748592377)
(2, 0.040734268724918366)
(3, 0.044923316687345504)
(4, 0.04808418080210686)
(5, 0.05311977490782738)
(6, 0.05407049134373665)
(7, 0.04193510487675667)
(8, 0.04387806728482246)
(9, 0.05169888958334923)
(10, 0.05626068636775017)
(11, 0.05652536451816559)
(12, 0.08336323499679566)
(13, 0.11564494669437408)
(14, 0.15328440070152282)
(15, 0.16894599795341492)
(16, 0.1790658235549927)
(17, 0.17483703792095185)
(18, 0.17168602347373962)
(19, 0.14102803170681002)
(20, 0.08823288977146148)
(21, 0.061944324523210526)
(22, 0.01280222274363041)
(23, 0.01035307813435793)
(24, 0.08275964111089705)

        };
    \addlegendentry{ASR}

    \addplot[smooth,color=green!45!black]
        plot coordinates {
(0, 5.906312026127125e-07)
(1, 0.00013600169950223062)
(2, 0.0001646773014847318)
(3, 0.0005568868794054162)
(4, 0.00016158985050904202)
(5, 0.0009009218752321212)
(6, 0.0006251518792131511)
(7, 0.0002711293816269407)
(8, 0.00010889265117195469)
(9, 3.9738095186895767e-06)
(10, 3.4027788297329247e-06)
(11, 2.844644086510819e-06)
(12, 3.769202066902237e-06)
(13, 0.00014969207313697552)
(14, 7.953346994327148e-07)
(15, 5.230138413102597e-05)
(16, 0.00035236583789811055)
(17, 0.02271812226622963)
(18, 0.027062036902961148)
(19, 0.08697226927223206)
(20, 0.018664656220696546)
(21, 0.039915796370314195)
(22, 5.740849585933396e-07)
(23, 5.262983382436913e-10)
(24, 0.80145581)

        };
    \addlegendentry{SID}
    \end{axis}
    \end{tikzpicture}

    \begin{tikzpicture}[thick,scale=0.78, every node/.style={scale=1}]
    \begin{axis}[
        legend style={at={(0.12,0.71)},anchor=center},
        title style={at={(0.5,0.97)},anchor=north, yshift=-0.1, draw=gray},
        xlabel=Layers (WavLM),
        ylabel=Weights,
        ymajorgrids, tick align=inside,
        major grid style=dashed,
        height=5cm, 
        width=10cm,
        xmin=0, xmax=25,
        ymin=0, ymax=1.0,
        ylabel near ticks,
        ytick={      0.1,  0.2, 0.3,  0.4,  0.5, 0.6, 0.7, 0.8, 0.9, 1.0, 1.1},
        yticklabels={0.05, 0.1, 0.15, 0.2, $//$, 0.7, 0.8, 0.9, 1.0}
        ]
    \addplot[smooth,color=blue] plot coordinates {
(0, 0.05180855189127922)
(1, 0.05269655585290014)
(2, 0.05093583446242498)
(3, 0.05164115130901337)
(4, 0.05071339901192322)
(5, 0.05112603365054131)
(6, 0.05183502286672592)
(7, 0.051041245735219954)
(8, 0.050746059659517594)
(9, 0.05087025836110715)
(10, 0.050975326448718906)
(11, 0.05103119090194747)
(12, 0.05521820593952333)
(13, 0.05476327434066772)
(14, 0.05835961177945137)
(15, 0.05621982365846634)
(16, 0.06357234716154405)
(17, 0.09132279455631774)
(18, 0.1112945324230194)
(19, 0.13878891918207917)
(20, 0.15984243130877556)
(21, 0.17282435297960003)
(22, 0.18216455970299867)
(23, 0.18937819411071566)
(24, 0.049027535339096065)
    };
    \addlegendentry{IC}

    \addplot[smooth,color=red]
        plot coordinates {
(0, 0.07792238146066666)
(1, 0.07793823629617691)
(2, 0.07789063453674315)
(3, 0.07786191254854202)
(4, 0.07788093388080597)
(5, 0.07787979394197464)
(6, 0.07788006216287613)
(7, 0.07792097330093384)
(8, 0.07798333466053009)
(9, 0.07798579335212708)
(10, 0.07789085060358048)
(11, 0.07795657962560654)
(12, 0.07812432199716568)
(13, 0.07901839166879654)
(14, 0.08137310296297074)
(15, 0.08262285590171814)
(16, 0.08263992518186569)
(17, 0.0830492228269577)
(18, 0.08288905769586563)
(19, 0.08324032276868819)
(20, 0.08356204628944397)
(21, 0.08350327610969544)
(22, 0.08351610600948335)
(23, 0.08358172327280044)
(24, 0.077888123691082)
        };
    \addlegendentry{KS}

    \addplot[smooth,color=cyan]
        plot coordinates {
(0, 0.07994253933429718)
(1, 0.06275319308042526)
(2, 0.055314723402261734)
(3, 0.047642625868320466)
(4, 0.04856565594673157)
(5, 0.04547930881381035)
(6, 0.03701106831431389)
(7, 0.03230048343539238)
(8, 0.03379451110959053)
(9, 0.05688457563519478)
(10, 0.05829453840851784)
(11, 0.036827538162469864)
(12, 0.05385073274374008)
(13, 0.07226462662220001)
(14, 0.06409107893705369)
(15, 0.06488530337810516)
(16, 0.09343890100717545)
(17, 0.10880437493324281)
(18, 0.09331819415092468)
(19, 0.12679269909858704)
(20, 0.15579059720039367)
(21, 0.1910857856273651)
(22, 0.10447507351636887)
(23, 0.10032641142606735)
(24, 0.17606541514396668)
        };
    \addlegendentry{ASR}

    \addplot[smooth,color=green!45!black]
        plot coordinates {
(0, 0.00016090939640909197)
(1, 0.00016360593484781887)
(2, 0.0005381155642680823)
(3, 0.006165734957903625)
(4, 0.005287696768962307)
(5, 0.0063519203104078766)
(6, 0.18734338878052666)
(7, 0.06689000874747766)
(8, 0.011356526624241074)
(9, 0.13766411456729888)
(10, 0.05374501644557503)
(11, 0.043618066437371615)
(12, 0.0229315124521014)
(13, 0.01062457915395498)
(14, 0.00022632286136390663)
(15, 0.00028353885663907637)
(16, 0.00016010469542970425)
(17, 0.00011490674914291465)
(18, 0.00014901705918585817)
(19, 0.00012417222071258543)
(20, 0.0002294642780796489)
(21, 0.0013688057537970136)
(22, 0.006351920766946601)
(23, 0.0013321095940597354)
(24, 0.61490571)
        };
    \addlegendentry{SID}

    \end{axis}
    \end{tikzpicture}

\begin{tikzpicture}[thick,scale=0.78, every node/.style={scale=1}]
    \begin{axis}[
        legend style={at={(0.14,0.65)},anchor=center},
        title style={at={(0.5,0.97)},anchor=north, yshift=-0.1, draw=gray},
        xlabel=Layers (Weighted-sum \textsc{Medium}),
        ylabel=Weights,
        ymajorgrids, tick align=inside,
        major grid style=dashed,
        height=5cm, 
        width=10cm,
        xmin=0, xmax=25,
        ymin=0, ymax=1.3,
        ylabel near ticks,
        ytick={      0.1,  0.2, 0.3,  0.4, 0.5,  0.6, 0.7, 0.8, 0.9, 1.0, 1.1, 1.2},
        yticklabels={0.05, 0.1, 0.15, 0.2, 0.25, 0.3, 0.35, 0.4, 0.45, 0.5, $//$, 1.0}
        ]
    \addplot[smooth,color=blue] plot coordinates {
(0, 0.044503916054964066)
(1, 0.042486388236284256)
(2, 0.04189660772681236)
(3, 0.042028382420539856)
(4, 0.04324157536029816)
(5, 0.042266424745321274)
(6, 0.042046066373586654)
(7, 0.04190237075090408)
(8, 0.0420348197221756)
(9, 0.04244065657258034)
(10, 0.04354889318346977)
(11, 0.04359303042292595)
(12, 0.04320519417524338)
(13, 0.05637393146753311)
(14, 0.07111570239067116)
(15, 0.09130856394767762)
(16, 0.1096690222620964)
(17, 0.12846916913986205)
(18, 0.1399613916879332)
(19, 0.14254212379455567)
(20, 0.13610130548477173)
(21, 0.14000762999172056)
(22, 0.1455844193696976)
(23, 0.14269977807998657)
(24, 0.14097262918949128)
    };
    \addlegendentry{IC}

    \addplot[smooth,color=red]
        plot coordinates {
(0, 0.07401218265295028)
(1, 0.07382470369338989)
(2, 0.07374200969934464)
(3, 0.07374068349599838)
(4, 0.07369847595691681)
(5, 0.0736640989780426)
(6, 0.07356838136911393)
(7, 0.07352610677480698)
(8, 0.07347942143678665)
(9, 0.0734485536813736)
(10, 0.07339860498905182)
(11, 0.07324381917715072)
(12, 0.07303674519062042)
(13, 0.08489415794610977)
(14, 0.08588019758462906)
(15, 0.08653116971254349)
(16, 0.08693914860486984)
(17, 0.08702930808067322)
(18, 0.08732858300209046)
(19, 0.08738750964403152)
(20, 0.08747612684965134)
(21, 0.08748960494995117)
(22, 0.0875292718410492)
(23, 0.08751198066568375)
(24, 0.0876200869679451)
        };
    \addlegendentry{KS}

    \addplot[smooth,color=cyan]
        plot coordinates {
(0, 0.017154743894934654)
(1, 0.004705052822828293)
(2, 0.003782296320423484)
(3, 0.003460074309259653)
(4, 0.0031273539643734693)
(5, 0.0035840626806020736)
(6, 0.0034103156067430974)
(7, 0.003002460347488523)
(8, 0.0034455957356840372)
(9, 0.002956074895337224)
(10, 0.0024926832411438226)
(11, 0.002215620130300522)
(12, 0.003044213866814971)
(13, 0.012377550825476645)
(14, 0.015024684369564056)
(15, 0.015449129976332188)
(16, 0.00852407794445753)
(17, 0.15335765480995178)
(18, 0.21206475794315337)
(19, 0.11021848767995834)
(20, 0.04208040982484818)
(21, 0.09731650352478028)
(22, 0.16412995755672453)
(23, 0.06824704259634018)
(24, 1.0448291301727294)
        };
    \addlegendentry{ASR}

    \addplot[smooth,color=green!45!black]
        plot coordinates {
(0, 0.0016322407172992826)
(1, 0.0013944520615041256)
(2, 0.0014346884563565253)
(3, 0.0014800244243815542)
(4, 0.0016476456075906754)
(5, 0.0020916203502565623)
(6, 0.005543064326047897)
(7, 0.018708476796746254)
(8, 0.011688516475260258)
(9, 0.015889614820480347)
(10, 0.06845653802156448)
(11, 0.25948342680931093)
(12, 0.5814005136489868)
(13, 0.07504550367593765)
(14, 0.7599910497665405)
(15, 0.06572355329990387)
(16, 0.08193135261535644)
(17, 0.024759616702795028)
(18, 0.011008130386471748)
(19, 0.0029332218691706658)
(20, 0.001844517071731389)
(21, 0.0015355173964043392)
(22, 0.001501779421232641)
(23, 0.0014925743453204632)
(24, 0.0013823382323607801)
        };
    \addlegendentry{SID}

    \addplot[smooth,color=violet,mark=x]
        plot coordinates {

(24, 1.2)
        };
    \addlegendentry{Vanilla}

    \end{axis}
    \end{tikzpicture}

\caption{The weight coefficients distribution of layers. The x-axis denotes different layers; the y-axis denotes the weight coefficients. Vanilla: the weight of vanilla {Whisper} on tasks.}
\label{fig:layer-info-plot}
\end{figure}
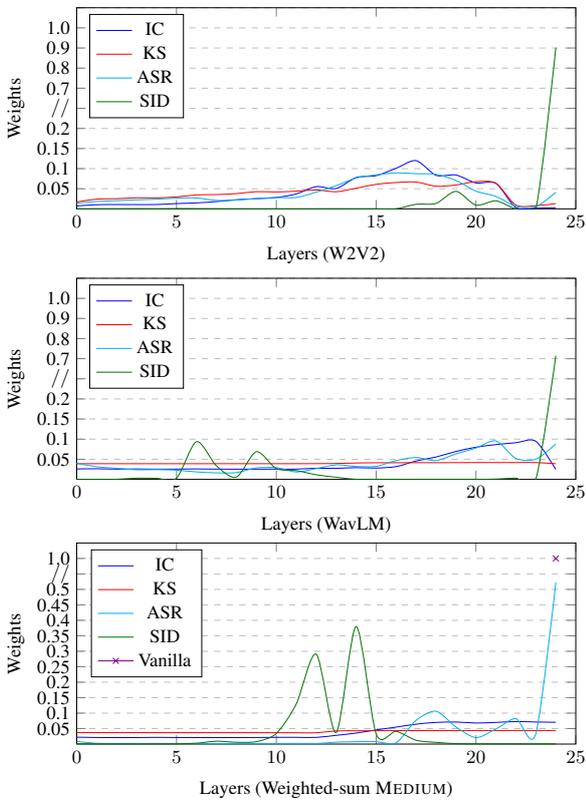
\begin{table*}[t]
\caption{Task performance of {Whisper} models, under different settings of task fine-tuning: Vanilla (V) denotes the Whisper encoder is frozen and its last-layer output is used as the task feature. Weighted-sum (W) is similar to Vanilla except for the construction of task features for which the weighted-sum of each layer's hidden state is taken. Fine-tuned {Whisper} (F) is also similar to Vanilla, except the encoder is no longer frozen and is fine-tuned together with the downstream task. -: The Whisper \textsc{Medium} could not be fine-tuned due to the limitation of GPU memory.}
\label{tab:others}
    \setlength{\tabcolsep}{10pt}
    \centering
    \scalebox{0.83}{
    \begin{tabular}{l c c c c c c c c c c c c c}
        \toprule
         & &\multicolumn{3}{c}{KS$\uparrow$}&\multicolumn{3}{c}{IC$\uparrow$}&\multicolumn{3}{c}{ASR$\downarrow$}&\multicolumn{3}{c}{SID$\uparrow$} \\
         \cmidrule(lr){3-5}\cmidrule(lr){6-8}\cmidrule(lr){9-11}\cmidrule(lr){12-14}
         Tr.  &Model  &V&W&F&V&W&F&V&W&F&V&W&F\\
         \cmidrule(lr){1-14}
    \bf \parbox[t]{2mm}{\multirow{3}{*}{\rotatebox[origin=c]{90}{$1\%$}}} 
    & \textsc{Base}  &   \textbf{96.79} & 94.87 & 93.12 & \textbf{67.04} & 34.06 & 41.92 & \textbf{26.43} & 31.97 & 62.00 &2.66 & \textbf{6.35} & 5.61   \\
    & \textsc{Small}  & 96.62 & 95.81 & \textbf{97.31} & 57.63 & 33.54 & \textbf{72.24} & \textbf{20.27} & 25.71 & 44.68 & 3.35 & \textbf{8.71} & 7.32   \\
    & \textsc{Medium}  & \textbf{96.72} & 95.52 & - & \textbf{73.74} & 34.19 & - & \textbf{17.56} & 24.56 & - & 3.97 & \textbf{12.48} & -   \\
    
    \midrule
    \bf \parbox[t]{2mm}{\multirow{3}{*}{\rotatebox[origin=c]{90}{$5\%$}}} 
    & \textsc{Base}  &   \textbf{97.44} & 97.31 & 95.52 & 95.39 & 90.46 & \textbf{96.28} & \textbf{16.18} & 17.37 & 30.51 & 11.63     & 25.74 & \textbf{25.90}\\
    & \textsc{Small}  &   \textbf{97.73} & 97.31 & 97.05 & 95.78 & 90.46 & \textbf{95.86} & \textbf{11.76} & 12.85 & 41.61  &13.47 & \textbf{36.90} & 34.07  \\
    & \textsc{Medium}  & \textbf{97.95} & 97.57 & - & \textbf{98.23} & 89.48 & - & \textbf{9.75} & 11.12 & - & 17.94 & \textbf{45.22} & -   \\
    
    \midrule
    \bf \parbox[t]{2mm}{\multirow{3}{*}{\rotatebox[origin=c]{90}{$10\%$}}} 
    & \textsc{Base}  &   \textbf{97.24} & \textbf{97.24} & 95.88 & 96.92 & 90.43 & \textbf{97.47} & \textbf{13.41} & 13.59 & 25.11 & 19.48 & 41.20  & \textbf{45.78}   \\
    & \textsc{Small}  &   \textbf{97.63} & 97.53 & 97.37 & 96.44 & 93.33 & \textbf{98.10} & \textbf{9.47} & 10.09 & 38.03  & 23.04 & \textbf{55.02} &  53.55  \\
    & \textsc{Medium}  & \textbf{97.96} & 97.54 & - & \textbf{98.78} & 95.28 & - & \textbf{7.74} & 8.52 & - & 30.05 & \textbf{65.51} & -   \\
   
    \bottomrule
    \end{tabular}
}
\end{table*}
\newline
 %
\noindent\textbf{Task Fine-tuning.} {
We adopt different strategies to generate speech representations. 
For the baselines, the weighted-sum of hidden states of each layer is considered as the feature for downstream task heads. For {Whisper} models, we considered only the last-layer output from the encoder\footnote{The pre-training process of {Whisper} involves an encoder-decoder where the encoder provides the last-layer output to its decoder, but for the other baselines (which are encoder-only models), more emphasis is placed on the connection between the intermediate encoder layers, suggesting a higher gain for them to be achieved from aggregating information across several layers for downstream tasks.} as speech representation for SUPERB downstream fine-tuning. We refer to this as the \emph{Vanilla} configuration, which is our default in the previous sections. In Table~\ref{tab:others}, we compare the performance of the \emph{Vanilla} setting with that of the \emph{Weighted-sum} and \emph{Fine-tuned} Whisper.\footnote{{We freeze Whisper for \emph{Weighted-sum} and unfreeze for \emph{Fine-tuned}.}}
We observe that across most tasks the \emph{Vanilla} version works best, while for the SID task, the \emph{Weighted-sum} representations is a better choice  suggesting that speaker information is retained in the intermediate layers of Whisper.
{In general, the \emph{Fine-tuned} Whisper that uses the last-layer output as features underperforms the \emph{Vanilla} variant with the frozen encoder, in the extreme low-data conditions (i.e., 1\%). We speculate this occurs as the captured knowledge stored in Whisper will be disrupted  after fine tuning the model on the small task data. As the size of training corpus increases, the trend continues for KS and ASR. For IC and SID, depending on the size of Whisper, \emph{Fine-tuning} may surpass \emph{Vanilla}.}

\noindent\textbf{Weight Coefficients Distribution.} We visualise the distribution of the weight coefficients (i.e., signifying the contribution of the corresponding layer in task fine-tuning) attached to each layer of Transformer and learned during the fine-tuning step in Figure~\ref{fig:layer-info-plot}. The numbers are based on fine-tuning in the 10\% training size. For Wav2vec2, WavLM and Whisper \textsc{Medium} encoders on 4 tasks we observe different patterns of layer contribution layers. The Figure also reveal that the speech features are distributed in various layers of the encoders. An interesting pattern is SID which places more emphasis on the last layers of W2V2 and WavLM, but shifts that to the intermediate layers of Whisper, indicating a stronger presencce of speaker features in its intermediate layers. As expected we observe that for Whisper and ASR task, most of the importance is placed on the final layers of the encoder. This also verifies why the \emph{Vanilla} configuration (which uses the last-layer as the task feature) is better at content-related tasks compared with a speaker task like SID. 
\noindent\textbf{Summary.}
We highlighted the quality of speech representations generated by Whisper. Compared to W2V2 and WavLM, the Whisper \textsc{Base} and \textsc{Small} have notably much fewer parameters, less than 100M, which leads to faster training convergence and inference. The representations achieve state-of-art performance on several downstream tasks. Regarding specific tasks (refer to Table~\ref{tab:others}), we observed that the \emph{Vanilla} Whisper for content tasks (ASR and KS) performs well in the very low-resource scenario. 
On IC, we observed that {Whisper} benefits more from fine-tuning with increasing number of training instances and model size. \emph{Vanilla} did not do well overall on SID whereas the \emph{Weighted-sum} improved the results substantially. 


\section{Conclusion}
In this paper, we evaluated the performance of three widely used pre-trained speech encoders in the low-resource setting on 7 diverse speech tasks from the SUPERB and SUPERB-SG benchmarks. We analysed the generated speech representations, for their qualitative and quantitative properties. Additionally, we looked at the internal contribution of layers from these encoders in various downstream task settings. Our findings highlighted the superior capabilities of the recent Whisper model's encoder for most of the semantic-content tasks and its performance degradation on speaker-focused task. We established a connection between the pre-training protocol of these models and their representational properties, and their downstream task performance.




\bibliographystyle{IEEEtran}
\bibliography{Interspeech2023}

\end{document}